\documentclass[sigconf]{acmart}
\usepackage{algorithm}
\usepackage{algorithmic}
\usepackage{dcolumn,booktabs}
\usepackage{multirow}
\usepackage[flushleft]{threeparttable}
\usepackage{subfigure}
\usepackage{amsmath}
\usepackage{amsfonts}
\usepackage{bm}
\usepackage{dsfont}
\usepackage{makecell}
\usepackage{enumitem}

\AtBeginDocument{%
  \providecommand\BibTeX{{%
    \normalfont B\kern-0.5em{\scshape i\kern-0.25em b}\kern-0.8em\TeX}}}

\setcopyright{acmcopyright}
\copyrightyear{2023}
\acmYear{2023}
\setcopyright{acmlicensed}\acmConference[WWW '23]{Proceedings of the ACM Web Conference 2023}{May 1--5, 2023}{Austin, TX, USA}
\acmBooktitle{Proceedings of the ACM Web Conference 2023 (WWW '23), May 1--5, 2023, Austin, TX, USA}
\acmPrice{15.00}
\acmDOI{10.1145/3543507.3583382}
\acmISBN{978-1-4503-9416-1/23/04}

\begin{document}

\title{DIWIFT: Discovering Instance-wise Influential Features for Tabular Data}

\author{Dugang Liu}
\additionalaffiliation{%
  \institution{Guangdong Laboratory of Artificial Intelligence and Digital Economy (SZ)}
  \city{Shenzhen}
  \country{China}
}
\authornote{Equal contribution}
\affiliation{%
  \institution{CSSE, Shenzhen University}
  \city{Shenzhen}
  \country{China}}
\email{dugang.ldg@gmail.com}

\author{Pengxiang Cheng}
\authornotemark[2]
\author{Hong Zhu}
\authornotemark[2]
\affiliation{%
  \institution{Huawei Technologies Co Ltd}
  \city{Shenzhen}
  \country{China}}

\author{Xing Tang}
\authornote{This work was done when working at Huawei Technologies Co Ltd}
\authornotemark[2]
\affiliation{%
  \institution{Tencent, FIT}
  \city{Shenzhen}
  \country{China}}
\email{shawntang@tencent.com}

\author{Yanyu Chen}
\author{Xiaoting Wang}
\affiliation{%
  \institution{Tsinghua-Berkeley Shenzhen Institute}
  \city{Shenzhen}
  \country{China}}

\author{Weike Pan}
\authornote{Corresponding authors}
\author{Zhong Ming}
\authornotemark[4]
\affiliation{%
  \institution{CSSE, Shenzhen University}
  \city{Shenzhen}
  \country{China}}

\author{Xiuqiang He}
\authornotemark[3]
\affiliation{%
  \institution{Tencent, FIT}
  \city{Shenzhen}
  \country{China}}
\email{xiuqianghe@tencent.com}

\renewcommand{\shortauthors}{Liu, et al.}


\begin{abstract}
Tabular data is one of the most common data storage formats behind many real-world web applications such as retail, banking, and e-commerce. The success of these web applications largely depends on the ability of the employed machine learning model to accurately distinguish influential features from all the predetermined features in tabular data. Intuitively, in practical business scenarios, different instances should correspond to different sets of influential features, and the set of influential features of the same instance may vary in different scenarios. However, most existing methods focus on global feature selection assuming that all instances have the same set of influential features, and few methods considering instance-wise feature selection ignore the variability of influential features in different scenarios. In this paper, we first introduce a new perspective based on the influence function for instance-wise feature selection, and give some corresponding theoretical insights, the core of which is to use the influence function as an indicator to measure the importance of an instance-wise feature. We then propose a new solution for discovering instance-wise influential features in tabular data (DIWIFT), where a self-attention network is used as a feature selection model and the value of the corresponding influence function is used as an optimization objective to guide the model. Benefiting from the advantage of the influence function, i.e., its computation does not depend on a specific architecture and can also take into account the data distribution in different scenarios, our DIWIFT has better flexibility and robustness. Finally, we conduct extensive experiments on both synthetic and real-world datasets to validate the effectiveness of our DIWIFT.
\end{abstract}

\ccsdesc[500]{Computing methodologies~Feature selection}
\ccsdesc[500]{Information systems~Data mining}

\keywords{Tabular data, Instance-wise feature selection, Influence function, Self-attention network}



\maketitle

\section{Introduction}\label{sec:introduction}
Tabular data is one of the most common data storage formats prepared for modeling in many practical web applications, such as e-commerce~\cite{zhang2019deep}, fraud detection~\cite{dal2014learned} and anomaly detection~\cite{pang2021deep}.
Typically, in tabular data, each row represents an instance and each column represents a feature. The value in a table cell is the specific value for that feature in a data instance~\cite{doleschal2018chisel}. 
Note that in addition to the column of features, there may be a column of labels indicating the category to which the corresponding instance belongs, e.g., in a supervised task.
Unlike homogeneous data such as image, text or speech data, which often have strong spatial, semantic or temporal correlations, tabular data tends to be heterogeneous, and the correlation between different columns (or features) may be weak.
As an example, in Table~\ref{tab:example}, we present a slice of the UCI-bank\footnote{https://archive.ics.uci.edu/ml/datasets/bank+marketing} dataset used in the experiments, where ``age (n)'' and ``education (c)'' are two different columns and the correlation between them is ambiguous.
Obviously, this property makes it much harder to customize a model to get good results from tabular data than from homogeneous data.

\begin{table}[htbp]
\caption{An example of tabular data in the UCI-bank dataset. The parenthesized letter in each column name indicates the type of feature, where `n' is a numerical dense feature and `c' is a categorical sparse feature.}
\centering
\scalebox{0.73}{
\begin{tabular}{cccccc}
\specialrule{0.1em}{3pt}{3pt}
\textbf{Age (n)} & \textbf{Job (c)} & \textbf{Marital (c)} & \textbf{Education (c)} & \textbf{Balance (n)} & \textbf{Housing (c)} \\
\specialrule{0.05em}{3pt}{3pt}
30 & unemployed & married & primary & 1787 & no \\
33 & services & married & secondary & 4789 & yes \\
35 & management & single & tertiary & 1350 & yes \\
59 & blue-collar & married & secondary & 0 & yes \\
35 & management & single & tertiary & 747 & no \\
\specialrule{0.1em}{3pt}{3pt}
\end{tabular}}
\label{tab:example}
\end{table}

To perform a more efficient learning process in tabular data and be successful in some corresponding business applications, an important approach is select features in a targeted manner following the optimization objective of the custom model~\cite{arik2021tabnet}, i.e., to identify the most influential features from all the predetermined features. 
Intuitively, the necessity of feature selection for tabular data is mainly reflected in the following aspects:
1) \textit{Efficiency}. Tabular data usually consists of many dense numerical features and high-dimensional sparse features, which consume a lot of resources in the training and inference stages of the model. Selecting only the most influential features and feeding them into the model can greatly reduce the costs.
2) \textit{Accuracy}. The correlation between features in tabular data is ambiguous. Moreover, there are often many irrelevant or redundant features, which are difficult to be distinguished in advance in the feature pre-determining stage. Removing features that are not influential (i.e., irrelevant or redundant) will benefit the learning of a model. In particular, since each instance usually has a different set of influential features that are beneficial to a particular task, it may be more beneficial to keep different influential features in different instances.
3) \textit{Interpretability}. A good interpretability of the adopted model is usually expected in practical business applications. For example, in a credit card approval scenario with the use of an auxiliary model, we would expect the model to simultaneously provide some key factors that influence the decision. Identifying the most influential features helps to indicate the importance of each feature and enhances the interpretability of the results.

However, although many research works on learning from tabular data have been proposed, few of them focus on solving the problem of feature selection in tabular data~\cite{borisov2021deep}.
In particular, these research works can be mainly divided into two categories according to their problems and goals, including how to effectively model tabular data~\cite{luo2020network,katzir2020net,arik2021tabnet,ucar2021subtab}, especially leveraging neural networks, and how to capture feature interactions for tabular data~\cite{liu2020dnn2lr,xie2021fives,luo2019autocross,chen2022danets}.
In addition, some of the above works may have an implicit feature selection step in the modeling process, but most of them may suffer from the redundant features as they are not designed for the goal of feature selection.
On the other hand, in order to reduce the huge workload of manually identifying the most influential features from tabular data, existing works aiming at solving the problem of feature selection mostly focus on global feature selection~\cite{guyon2003introduction}, i.e., the granularity of selection is an entire column in tabular data and all instances have the same set of influential features.
The limitation of global feature selection is that in practice, the influential features w.r.t. different instances or a same instance in different environments may be different.
For this reason, we are motivated to design a robust instance-wise feature selection method for tabular data.

To the best of our knowledge, there is still a lack of research on instance-wise feature selection for tabular data compared with those for homogeneous data~\cite{chen2018learning,panda2021instance,yoon2018invase}. 
In this paper, we first introduce a new perspective based on the influence function for instance-wise feature selection. 
Moreover, we give some theoretical insights, i.e., how to use the influence function as an indicator to measure the importance of an instance-wise feature, so as to guide instance-wise feature selection.
We then propose a new solution for discovering instance-wise influential features in tabular data (DIWIFT).
Specifically, our DIWIFT mainly include a feature selection module with a self-attention network and a calculator for computing the value of the corresponding influence function, which will be used to guide the selection of some instance-wise features.
Our DIWIFT is of better flexibility and robustness due to the merits of the influence function, i.e., its computation does not depend on a specific architecture and it enables the model to trade off between training and validation distributions.
Finally, we conduct extensive experiments on three synthetic and four real-world datasets, where the results clearly the effectiveness and robustness of our DIWIFT.

\section{Related Work} \label{sec:related}
In this section, we briefly review some related works on three research topics, including tabular data modeling, feature selection and influence function.

\noindent\textbf{Tabular Data Modeling.}
Existing works on tabular data modeling can be mainly divided into two categories. 
The first category focuses on how to model tabular data more effectively~\cite{luo2020network,katzir2020net,arik2021tabnet,ucar2021subtab}, especially using neural networks.
For example, introducing more complex network structures to learn fusion of different features and increasing interpretability~\cite{luo2020network,arik2021tabnet}, or modeling tabular data through some new perspectives such as multi-view representation learning~\cite{ucar2021subtab}.
The second category aims to design some more efficient ways to capture feature interactions in tabular data modeling~\cite{liu2020dnn2lr,xie2021fives,luo2019autocross,chen2022danets}.
Unlike them, our DIWIFT focuses on addressing instance-wise feature selection in tabular data, which is rarely studied in existing works.
Note that some related works may also include feature selection as an incidental output, such as TabNet~\cite{arik2021tabnet}, but since feature selection is not their main goal, they may still suffer from feature redundancy.
In addition, our DIWIFT can be used as a pre-feature selection module to integrate with these tabular data modeling methods to enhance their effectiveness and efficiency.

\noindent\textbf{Feature Selection.}
Feature selection often refers to discovering a subset of features based on their usefulness.
Most existing methods focus on global feature selection, where the importance of each feature is assigned based on the entire training data~\cite{guyon2003introduction}.
To achieve instance-wise feature selection, many previous studies on homogeneous data (instead of on tabular data) have been proposed, where the number of influential features per instance is assumed to be the same and different, respectively~\cite{chen2018learning,panda2021instance,yoon2018invase}.
Note that feature selection may still be beneficial in deep models in addition to traditional machine learning models~\cite{lyu2023feature,lyu2023optimizing}.
However, feature selection is generally not a major optimization goal in existing works on tabular data modeling, and there are very few works on instance-wise feature selection in tabular data.
Our DIWIFT aims to bridge the gap in this research direction.
In addition, our DIWIFT is also easy to integrate with existing feature selection methods by using the proposed influence function-based loss as their auxiliary loss to improve the performance of feature selection.

\noindent\textbf{Influence Function.}
Influence function (IF) is an important concept in the scope of robust statistics and is defined by the Gateaux derivative~\cite{huber2011robust}.
It can be used to measure instance-wise influence~\cite{koh2017understanding,yu2020influence,ren2018learning,wang2020less} and feature-wise influence~\cite{sliwinski2019axiomatic} on a validation loss.
These obtained influences can be used to construct a sampling strategy for the important instances~\cite{wang2020less,wang2018data}, or to reweight the biased training instances in an optimization objective~\cite{yu2020influence,ren2018learning}, etc.
We find that most of the previous works on IF focus on the instance level, and rarely involve the feature level as that in the studied problem of this paper.
Furthermore, there is no general and systematic analysis on how to guide the use of IF in feature selection.
Our DIWIFT is a novel IF-based instance-wise feature selection method for tabular data.

\section{Preliminaries} \label{sec:preliminaries}
In this paper, we focus on feature selection in supervised learning.
The training instances $\{z_i\}_{i=1}^{n}=\{(\bm{x}_i,y_i)\}_{i=1}^{n} \in \mathcal{X} \times \mathcal{Y}$ are drawn from a training distribution $P(\bm{x},y)$, where $n$ is the number of the training instances, $\mathcal{X}=\mathcal{X}_1\times\dots\times\mathcal{X}_d$ is the $d$-dimensional feature space, and $\mathcal{Y}$ is the discrete label space, which is $\{0,1\}$ in binary classification and $\{1,\dots,c\}$ in multi-class classification. 
A prediction model trained on a given training sample $\{z_i\}_{i=1}^{n}$ can be obtained by minimizing the empirical risk, i.e., $\hat{\theta} \triangleq \arg\min_{\theta\in\Theta} \frac1n \sum_{i=1}^n l(z_i,\theta)$.
Note that to simplify the notation, we omit the regularization term in the loss.
We put the main notations in Table~\ref{tab:notation} for ease of reference.

\begin{table}[htbp]
  \centering
  \caption{The main notations and their explanations.}
    \begin{tabular} {c|p{6.7cm}}
    \toprule
	Symbol   & Meaning \\
	\hline
    $z_i$, $z_j$   & $i$-th training and $j$-th validation instance. \\
    $i$, $j$, $k$    & Index of training instance, validation instance, and feature. \\
    $n$, $m$, $d$       & Size of training set, evaluation set, and feature dimension. \\
    $\delta_i$, $\delta_{ik}$     & Perturbation on features of $z_i$, $\delta_i\in\mathbb{R}^d$, $\delta_{ik}\in\mathbb{R}$ . \\
    $S$, $S_{ik}$   &   Feature selection matrix and its element for training set, $S\in\mathbb{R}^{n\times d}$, $S_{ik}\in\mathbb{R}$.\\
    $\hat{\theta}$, $\theta$  &  Parameter of base model.\\
    $\hat{\omega}$, $\omega$  &  Parameter of self-attention network.\\
    $\phi(z_i,z_j)$     & Influence function of $z_i$ on $z_j$, $\phi(z_i,z_j)\in\mathbb{R}^{1\times d}$. \\
    $\phi_i$, $\phi_{ik}$     & Influence function of $z_i$ on the whole validation set, $\phi_i\in\mathbb{R}^{1\times d}$, $\phi_{ik}\in\mathbb{R}$.  \\
    $l(\cdot)$     & Loss of base model, $l(\cdot)\in\mathbb{R}$.\\
    $P(\bm{x},y)$ &  Training distribution. \\
    $Q(\bm{x},y)$  &  Validation distribution. \\
    \bottomrule
    \end{tabular}%
  \label{tab:notation}%
\end{table}%

\subsection{Feature Selection Matrix in Training Data}\label{subsec:feature}
Instead of global feature selection which selects a same subset of features for all the instances, we consider a more complex case where different instances depend on different subsets of features, and then aim to improve the model performance through instance-wise feature selection.
We refer to $S\in\{0,1\}^{n\times d}$ as the feature selection matrix for the training set.
Note that the number $n$ of instances in a tabular data is often much larger than the number $d$ of columns (or features). 
Each row and column in the feature selection matrix $S$ corresponds to each instance and each feature in the training set, respectively.
Therefore, in the feature selection matrix $S$, it can be represented as 1 if a feature of an instance is preserved, and 0 otherwise.
To sum up, the meaning of $S$ is,
\begin{equation}
S_{ik}=
    \begin{cases}
    1 & \text{if feature } k \text{ is selected in }z_i, \\
    0 & \text{if feature } k \text{ is not selected or is zero in }z_i.
    \end{cases}
\end{equation}

\subsection{Definition of Influence Function}\label{subsec:definition}
How to measure the influence of a feature on model performance is the key question in feature selection, for which we utilize the influence function (IF). 
We first briefly introduce the definition of the feature-level influence function.
If a training instance $z_i$ is perturbed to $z_i'=(\bm{x_i}+\delta_i,y_i)$, the influence of the perturbation on the loss at a validation instance $z_j$ has a closed-form expression~\cite{koh2017understanding}:
\begin{equation} \label{equ:if}
\begin{split}
\phi(z_i, z_j)  
&\triangleq \frac{dl(z_j,\hat{\theta}_{\delta_i})}{d\delta_i} |_{\delta_i=0} \\
&=-\nabla_{\theta}l(z_j,\hat{\theta})^{\top}H^{-1}_{\hat{\theta}}\nabla_{\bm{x}}\nabla_{\theta}l(z_i,\hat{\theta}),
\end{split}
\end{equation}
where $\hat{\theta}_{\delta_i}$ is the empirical risk minimizer after $z_i$ is perturbed to $z_i'$, $\phi(z_i, z_j)\in \mathbb{R}^{1\times d}$ is the feature-level IF of $z_i$ over $z_j$, $z_j$ is a validation instance draw from a validation distribution $Q(\bm{x},y)$, and $H_{\hat{\theta}}\triangleq \frac1n \sum_{i=1}^n \nabla^2_{\theta}l(z_i,\theta)$ is a positive and definite Hessian matrix.
Note that previous studies targeting feature-level IF are few.
Different from the feature-level IF in Eq.\eqref{equ:if}, instance-level IF has been exploited in many previous studies on instance sampling and instance reweighting~\cite{yu2020influence,ren2018learning,wang2020less}.

\section{The Proposed Method}\label{sec:diwift}
In this section, we first introduce some theoretical insights on how to use the influence function to discover some most influential instance-wise influential features.
We then propose a new method for discovering instance-wise influential features in tabular data and describe it in detail.

\begin{figure*}[htbp]
    \centering
    \includegraphics[width=0.95\linewidth]{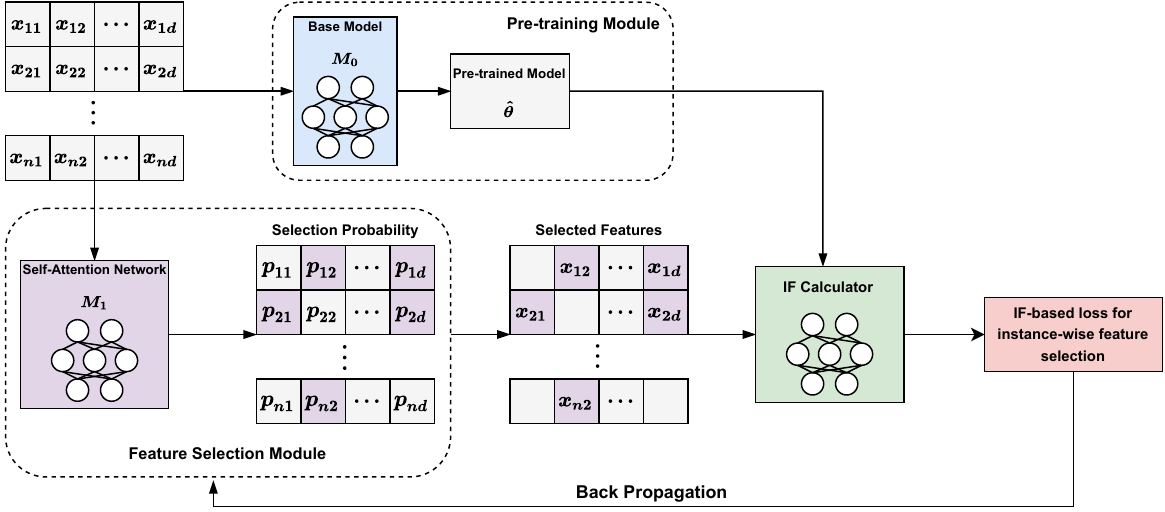}
    \caption{The overall architecture of the proposed DIWIFT, where the core components are a feature selection model with a self-attention network, and an IF calculator for calculating the value of the corresponding influence function.}
    \label{fig:framework}
\end{figure*}   

\subsection{Theoretical Analysis}\label{subsec:theoretical}
According to the definition of feature-level IF in Eq.\eqref{equ:if}, we can approximate the loss change of $z_j\sim Q(\bm{x},y)$ if $z_i\sim P(\bm{x},y)$ is perturbed by $\delta_i\in\mathbb{R}^d$,
\begin{equation}
    l(z_j,\hat{\theta}_{\delta_i})-l(z_j,\hat{\theta})\approx\phi(z_i,z_j) \delta_i.
\end{equation}
This process can be extended to the whole validation set as follows,
\begin{equation}\label{loss-change}
    \sum_{j=1}^{m} l(z_j,\hat{\theta}_{\delta_i})-\sum_{j=1}^{m} l(z_j,\hat{\theta}) \approx [\sum_{j=1}^{m}\phi(z_i,z_j)] \delta_i,
\end{equation}
where $m$ is the size of the validation set. 
Obviously, the best perturbation should minimize the validation loss $\sum_{j=1}^{m} l(z_j,\hat{\theta}_{\delta_i})$. We then have the optimization problem about $\delta_i$,
\begin{equation} \label{equ:optDelta}
\begin{aligned}
    \delta_{i}^* = \arg\min_{\delta_{i}} \sum_{j=1}^{m} l(z_j,\hat{\theta}_{\delta_i})
    =\arg\min_{\delta_{i}}
    \phi_{i} \delta_{i},
\end{aligned}
\end{equation}
where $\phi_i=\sum_{j=1}^{m}\phi(z_i,z_j)\in\mathbb{R}^{1\times d}$ indicates the influence of instance $z_i$ over the whole validation set. 

Let the $k$-th dimension of $\phi_i$, $\delta_{i}$ and $\bm{x}_i$ be denoted as $\phi_{ik}$, $\delta_{ik}$ and $\bm{x}_{ik}$, respectively. 
For the adversarial training problems where the feature values are dense and continuous, such as image data, the optimal $\delta_{i}^*$ is in the direction of $\phi_{i}^T$ \cite{koh2017understanding}. 
However, for the problem of feature selection in tabular data, the range of values for $\delta_{ik}$ is $\{0,-\bm{x}_{ik}\}$, where $\delta_{ik}=0$ means $\bm{x}_{ik}$ remains unchanged, and $\delta_{ik}=-\bm{x}_{ik}$ means $\bm{x}_{ik}$ is removed. 
Moreover, in tabular data, the one-hot encoded sparse feature value $\bm{x}_{ik}$ is in $\{0,1\}$, and the normalized dense feature value $\bm{x}_{ik}$ is in $[0,1]$. Thus, $\delta_{ik}\leq 0$ always holds, and the solution of Eq.\eqref{equ:optDelta} is,
\begin{equation}\label{equ:optimalDelta}
\delta_{ik}^*=
    \begin{cases}
    0 & \text{if } \phi_{ik} < 0, \\
    -\bm{x}_{ik} & \text{if } \phi_{ik} \geq 0.
    \end{cases}
\end{equation}
The results in Eq.\eqref{equ:optimalDelta} is intuitive because $\phi_{ik} \geq 0$ means the presence of $\bm{x}_{ik}$ will increase the validation loss, for which we should indeed remove this feature. 
Recall the feature selection matrix $S$ defined in Section \ref{sec:preliminaries}, we can see that, if a feature $k$ in $z_i$ is selected, i.e., $S_{ik}=1$, then both conditions should be satisfied, i.e., $\bm{x}_{ik}>0$ and $\delta_{ik}=0$. We then get the theoretically optimal $S^*_{ik}$,
\begin{equation} \label{equ:optimalS}
S_{ik}^*=\mathds{1}(\phi_{ik}\bm{x}_{ik} < 0),
\end{equation}
where $\mathds{1}(\cdot)$ is a 0-1 indicator function.

This means that by minimizing the validation loss of the model trained after instance-wise feature selection, we can obtain the optimal instance-wise feature selection strategy in Eq.\eqref{equ:optimalS} with the help of the influence function.
Note that the influence function allows the model to trade off between the training and validation distributions~\cite{yu2020influence}, and the resulting instance-wise feature selection method is expected to perform robustly in the scenarios where a distribution shift exist.
Next, we describe the proposed DIWIFT method in detail, which is a new instance-wise feature selection method based on the influence function, and verify its robustness in the experiments.
To the best of our knowledge, most existing works on feature selection have not considered the variability of influential features across different scenarios.

\subsection{Discovering Instance-wise Influential Features in Tabular Data}\label{methodSec}
In this section, based on the theoretical insights in Section~\ref{subsec:theoretical}, we propose a novel method for discovering instance-wise influential features in tabular data, or DIWIFT for short.

\subsubsection{Architecture}\label{subsubsec:architecture}
The overall framework of our DIWIFT is illustrated in Figure~\ref{fig:framework}.
As shown in Figure~\ref{fig:framework}, the core steps of our DIWIFT include: 
1) a pre-training module aims to train on a base model $M_0$ based on the original tabular data, i.e., without feature selection, in order to obtain a set of pre-trained model parameters $\hat{\theta}$. These model parameters are required in the subsequent steps to calculate the value of the influence function as shown in Eq.\eqref{equ:if}. Although $\hat{\theta}$ is theoretically the parameter of the optimal empirical risk minimizer, we may not be able to obtain an accurate $\hat{\theta}$ in practice because the deep learning model is non-convex. To this end, we will perform a sensitivity analysis about our DIWIFT with some pre-trained models of varying performance in our experiments;
2) a feature selection module with a self-attention network $M_1$ receives the original tabular data and outputs the corresponding instance-wise feature selection probabilities;
3) after feature selection is performed on each instance according to the selection probability, these new instances and the pre-trained model $\hat{\theta}$ are fed into the IF calculator to calculate the value of the corresponding influence function;
and 4) the calculated value of the influence function is used to calculate the IF-based loss designed in Eq.\eqref{loss} for instance-wise feature selection, and then the feature selection model $M_1$ is updated by means of back propagation.

Note that since the self-attention mechanism has shown its effectiveness and flexibility in previous related works~\cite{zhou2018din,skrlj2020feature,qin2020user}, we employ a self-attention network in the feature selection module as an example, to capture the importance of instance-wise features driven by the influence function.
However, the self-attention mechanism is not a necessary structure, and in fact, any neural network layer that can generate a mask matrix of the same dimension as the input can be used as a feature selection model.
After obtaining the feature selection model $M_1$, we need to refine the base model $M_0$ based on the original instance to obtain the final prediction model, where the original instance will first undergo a feature selection process through $M_1$.
Similarly, in the prediction stage, each instance will go through a process of instance-wise feature selection through $M_1$, and then get the predicted label that is fed into the final prediction model.
Next, we will give a detailed introduction to the important modules in our DIWIFT.

\subsubsection{Feature selection module} 
As described in Section~\ref{subsubsec:architecture}, we use a self-attention network in the feature selection module to effectively model the selection probability, since a self-attention~\cite{vaswani2017attention} has been proven to be a useful module that can capture important features in instances~\cite{zhou2018din,skrlj2020feature,qin2020user}.
We first use a multi-head attention on an instance to get its embedding representation.
Specifically, we have,

\begin{equation}\label{query}
    Q = K = V = \left(\mathbf{e}_{1};\mathbf{e}_{2};\dots;\mathbf{e}_{i};\dots;\mathbf{e}_{d}\right),
\end{equation}
where $\mathbf{e}_{i}$ denotes the embedding representation of the $i$-th feature, $Q = K = V  \in R^{d \times K_d}$ and $K_d$ is the dimension of the output embedding. 
Then, the computation of self-attention can be expressed as,
\begin{equation}\label{self_attention}
    Attention(Q,K,V) = softmax(\frac{QK^T}{\sqrt{K_d}})V.
\end{equation}
Further, the multi-head self-attention can be calculated as follows,
\begin{equation}
\begin{aligned}
    E &= Multihead(Q,K,V) \\
    &= Concatenate(head_1,head_2,\dots,head_h)W^O,
\end{aligned}
\end{equation}
where $h$ denotes the number of self-attention networks, $head_i$ = $Attention(Q_i,K_i,V_i)$ and $W^O \in R^{hK_d \times K_d}$ is a parameter matrix.
After obtaining the embedding representation $E$ of an instance, we feed it into a multi-layer perceptron (MLP) with a $ReLU$ activation function to obtain the corresponding output for each instance,
\begin{equation}
    f(x, \omega) = MLP(E),
\end{equation}
where $f(\bm{x},\omega):\mathbb{R}^d\rightarrow\mathbb{R}^d$ denotes the mapping from an input feature vector to a corresponding output, i.e., a feature selection model based on a self-attention network, and $\omega$ denotes the parameters of this model.

\begin{figure}[htbp]
    \centering
    \includegraphics[width=0.7\linewidth]{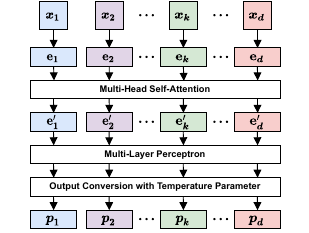}
    \caption{The structure of a feature selection model with a self-attention network, where the embedding representation of each instance is $E=\left(\mathbf{e}'_1,\mathbf{e}'_2,\dots,\mathbf{e}'_d\right)$.}
    \label{select}
\end{figure}

We design the probability of selecting each corresponding feature as
\begin{equation}\label{prob}
    p(\bm{x},\omega)=\sigma(f(\bm{x},\omega)/\tau),
\end{equation}
where $p(\bm{x},\omega)\in\mathbb{R}^d$, $\sigma(\cdot)$ is the sigmoid function, $\tau\in\mathbb{R}^+$ is the alterable temperature parameter. 
We propose to relate the temperature control coefficient to the number of training iterations, i.e., $\tau=\max(\tau_{min},1-(1-\tau_{min})t/t_{max})$, where $\tau_{min}$ is a sufficiently small minimum temperature parameter (e.g., 0.001 used in the experiments), $t$ is the current number of iterations, and $t_{max}$ is the maximum number of iterations.
Obviously, as the number of training iterations increases, $\tau$ will gradually decrease to a small enough value, and this will ensure that the selection probability of each feature is close to 0 or 1 to obtain the discrete feature selection mask.
For ease of understanding, we present a structure of the feature selection model in Figure~\ref{select}.

\subsubsection{IF calculator} 
In this subsection, we introduce the influence function to guide the training of a feature selection model and further improve the accuracy and robustness of instance-wise feature selection for tabular data.
According to the definition of IF in Eq.\eqref{equ:if}, we can find that the calculation of IF is usually complicated. 
Therefore, we need to solve the problem of how to calculate IF efficiently.
Let $p_i$ and $p_j$ be the selection probabilities of the features in $z_i$ and $z_j$, respectively, where $p_i=\left[p_{i1},p_{i2},\dots,p_{id}\right]$ and $p_j=\left[p_{j1},p_{j2},\dots,p_{jd}\right]$.
After obtaining the pre-trained model parameters $\hat{\theta}$ and reweighting the features using $p_i$ and $p_j$, the influence of $z_i$ on the entire validation set is,
\begin{equation}\label{phip}
\begin{aligned}
    \phi_i(p(\bm{x},\omega))=-[\sum_{j=1}^m\nabla_{\theta}l(p_j\odot\bm{x}_j,y_j,\hat{\theta})]^{\top} & \\ H_{\hat{\theta}}(p)^{-1}\nabla_{\bm{x}}\nabla_{\theta}l(p_i\odot\bm{x}_i,y_i,\hat{\theta}),
\end{aligned}
\end{equation}
where $\odot$ is the element-wise product and $H_{\hat{\theta}}(p)=\frac1n \sum_{i=1}^n \nabla^2_{\theta}l(p_i\odot\bm{x}_i,y_i,\hat{\theta})$. 
The influence function $\phi_i(p(\bm{x},\omega))$ can be calculated in three steps: 
1) we compute the \emph{inverse Hessian-vector-product} (HVP) $[\nabla_{\theta}\sum_{j=1}^ml(p_j\odot\bm{x}_j,y_j,\hat{\theta})]^{\top} H_{\hat{\theta}}(p)^{-1}$ and set the result as a constant vector independent of the derivative of $\bm{x}$;
2) for each training instance, we multiply the constant vector with $\nabla_{\theta}l(p_i\odot\bm{x}_i,y_i,\hat{\theta})$;
and 3) finally, we take the derivative with respect to $\bm{x}$.
The most difficult step is computing HVP, and we use a stochastic estimation method from previous studies~\cite{koh2017understanding} to efficiently handle high-dimensional and large-scale tabular data.
In the stochastic estimation method, let $H_u^{-1}=\sum_{v=0}^u(I-H)^v$ be the first $u$ terms in the Taylor expansion of $H^{-1}$, where $H$ is an arbitrary Hessian matrix and $I$ is the identity matrix.
We then have $H_u^{-1}=I+(I-H)H_{u-1}^{-1}$, $H_u^{-1}\rightarrow H^{-1}$ when $u\rightarrow \infty$\footnote{To ensure the validity of the Taylor expansion, $\forall i, \nabla^2_{\theta}l(z_i,\hat{\theta})\preccurlyeq I$ should be satisfied. This is always true because we can shrink the loss without affecting the parameters}.
The key step is that in each iteration, we can sample some instances to compute an unbiased estimator of $H$. 
Therefore, we can get the following calculation process for HVP: 
1) uniformly sample some instances from the training set and calculate the Hession matrix $\tilde{H}$;
2) define $H_0^{-1}\mu=\mu$, where $\mu=[\nabla_{\theta}\sum_{j=1}^ml(p_j\odot\bm{x}_j,y_j,\hat{\theta})]$ in the feature selection;
and 3) recursively compute $H_u^{-1}\mu=\mu+(I-\tilde{H})H_{u-1}^{-1}\mu$.
Note that the computational complexity of the original IF is $O(np^2 + p^3)$, where $p$ is the dimension of model parameters, and the complexity of stochastic estimation used is $O(np + rtp)$, where $r$ is the number of samples sampled, and $t$ is the number of sampling executions. 
We can find that DIWIFT is of good scalability by comparing these two complexities.
The complete process of calculating $\phi_i(p(\bm{x},\omega))$ is shown in Algorithm 1 of Appendix~\ref{appendix}.


\subsubsection{IF-based loss for instance-wise feature selection}
The final optimization objective of our DIWIFT is to minimize the sum of $\phi_{ik}$ of the selected $\bm{x}_{ik}$:
\begin{equation}\label{loss}
    \hat{\omega}=\arg\min_{\omega}\sum_{i=1}^{n}\phi_{i}(p(\bm{x},\omega))[p(\bm{x}_i,\omega)\odot\mathds{1}(\bm{x}_i)],
\end{equation}
where $\mathds{1}(\bm{x}_i) \in \mathbb{R}^d$ transfers the $k$-th dimension of $\bm{x}_i$ to 1 if $\bm{x}_{ik}>0$, and otherwise to 0. We can see that $p(\bm{x}_i,\omega)\odot\mathds{1}(\bm{x}_i)$ means feature $\bm{x}_{ik}$ has no probability to be selected if $\bm{x}_{ik}=0$. 
The complete training process of our DIWIFT is shown in Algorithm 2 of Appendix~\ref{appendix}.


\section{Empirical Evaluations}
In this section, we conduct experiments with the aim of answering the following five key questions. 
\begin{itemize}[leftmargin=*]
    \item Q1: How does our DIWIFT perform compared to the baselines?
    \item Q2: How well does our DIWIFT identify the instance-wise influential features?
    \item Q3: How do different modules of our DIWIFT contribute to its performance? 
    \item Q4: How does our DIWIFT perform in presence of distribution shift?
    \item Q5: How robust is our DIWIFT to fluctuations in a pre-trained model? 
\end{itemize}

\subsection{Experimental Setup}
\subsubsection{Datasets.}
To comprehensively evaluate the performance of our DIWIFT, we consider both some synthetic datasets and real-world datasets in our experiments.
We first generate three synthetic datasets following the approach adopted in previous related works~\cite{chen2018learning,yoon2018invase}.
Specifically, the input features are generated from an 11-dimensional Gaussian distribution, where there is no correlation between the features, i.e., $\bm{x} \sim \mathcal{N} (\bm{0}, \mathbf{I})$.
The $k$-th feature is denoted as $\bm{x}^k$.
The label $y$ is generated from a Bernoulli random variable with $\mathbb{P} (y=1|\bm{x})=\frac{1}{1+logit(\bm{x})}$, where $logit(\bm{x})$ can vary to create three different synthetic datasets:
\begin{itemize}
    \item \textbf{Syn1: } exp$\left( \bm{x}^1 \bm{x}^2 \right)$.
    \item \textbf{Syn2: } $-10 \times \sin 2\bm{x}^7 + 2\left |\bm{x}^8 \right | + \bm{x}^9 + \mathrm{exp} (-\bm{x}^{10})$.
\end{itemize}
In the above two datasets, the generation of labels $y$ depends on the same set of features for each instance.
To compare the ability of different methods to discover instance-wise influence features, by setting $\bm{x}^{11}$ as a switch feature, we create a new synthetic dataset, where different instances have different influence features.
\begin{itemize}
    \item \textbf{Syn3: } If $\bm{x}^{11} < 0$, $logit(\bm{x})$ follows \textbf{Syn1}; otherwise, $logit(\bm{x})$ follows \textbf{Syn2}.
\end{itemize}
In addition, we employ four datasets, including Coat~\cite{schnabel2016recommendations}, Adult~\cite{kohavi1996scaling}, Bank, and Credit~\cite{dal2014learned}, that are widely adopted in previous works focusing on modeling tabular data~\cite{kim2021oct,luo2019autocross}.
The statistics of all the datasets are summarized in Table~\ref{dataset}. In subsequent experiments, we divide all instances of each dataset into a training set, a validation set and a test set, where each part corresponds to a ratio of $3:1:1$.

\begin{table}[htbp]
  \centering
  \caption{Statistics of three synthetic datasets and four real-world datasets.}
    \scalebox{0.85}{
    \begin{tabular} {ccc}
    \toprule
     Datasets   &   \makecell[c]{\#Features} &  \#Instances \\
    \hline
    Syn1 &  11  &  30k \\
    Syn2 &  11  &  30k \\
    Syn3 &  11  &  30k \\
    Coat &  47 &  11k \\
    Adult &  137 &  49k \\
    Bank &  55  &  45k \\
    Credit &  30  &  284k \\
    \bottomrule
    \end{tabular}
    }
  \label{dataset}
\end{table}

\subsubsection{Baselines}
We choose the most representative methods from the two routes as the baselines, including two global feature selection methods, i.e., Lasso~\cite{fonti2017feature} and Tree~\cite{geurts2006extremely}, and four instance-wise feature selection methods, i.e., L2X~\cite{chen2018learning}, CL2X~\cite{panda2021instance}, INVASE~\cite{yoon2018invase} and TabNet~\cite{arik2021tabnet}.
\begin{itemize}[leftmargin=*]
\item Lasso~\cite{fonti2017feature}: it is a widely used global feature selection method via adding $L_1$ regularization to the loss of a linear model.
\item Tree~\cite{geurts2006extremely}: it is a global feature selection method via an extremely randomized trees classifier.
\item L2X~\cite{chen2018learning}: it is an instance-wise feature selection method that can discover a fixed number of influential features for each instance through mutual information. It is the first method to implement instance-wise feature selection and interpretation.
\item CL2X~\cite{panda2021instance}: it is a causal extension of L2X that also discovers a fixed number of influential features for each instance via conditional mutual information.
\item INVASE~\cite{yoon2018invase}: it is an instance-wise feature selection method that can discover an adaptive number of influential features for each instance by minimizing the Kullback-Leibler divergence between the full conditional distribution and a conditional distribution that includes only the selected set of features. It is an important baseline because it best matches the problem we focus on solving in this paper.
\item TabNet~\cite{arik2021tabnet}: it is an instance-based feature selection method that uses sequential attention to select the features that need to be inferred at each decision step. Furthermore, it proposes a novel high-performance and interpretable deep learning architecture for tabular data.
\end{itemize}

\subsubsection{Evaluation Metrics.}
We apply the Area Under the ROC Curve (AUC) as the evaluation metric, which is commonly adopted in previous studies of tabular data~\cite{xie2021fives,arik2021tabnet}. Specifically, the AUC in binary classification task is calculated by,
\begin{equation*}
    AUC=\frac{\sum_{z_p\in z^+, z_q\in z^-}\mathds{1}(g(z_p)>g(z_q))}{|z^+| |z^-|},
\end{equation*}
where $z^+$ is the set of positive instances, $z^-$ is the set of negative instances, $|z^+|$ and $|z^-|$ are their sizes; $z_p$ is a positive instance, $z_q$ is a negative instance; $g(\cdot)$ is a classifier.
Note that if the predicted scores of all positive samples are higher than the predicted scores of negative samples, the model will reach AUC$=1$ (perfect separation of positive/negative samples), i.e., the upper bound of AUC is 1, and the bigger the better.

\subsubsection{Implementation Details}
For all the methods, a three-layer MLP is adopted as a base model.
For the search range, $L_2$ regularization parameter is in $[1e^{-6}, 1]$, learning rate is in $[1e^{-5}, 1e-1]$, hidden layer size is in $\left\{50, 100, 150, 200 \right\}$, and batch size is in $\left\{64, 128, 256, 512, 1024, 2048 \right\}$.
A special $L_1$ parameter with LASSO is in $[1e^{-5}, 1e^{-1}]$, and Tree needs to set the number of trees from 3 to 30.
For L2X and CL2X , we set the number of features selected in each instance in the range of $[2,d]$. 
For our DIWIFT, temperature parameter is in $[1e^{-3}, 1]$.
Regarding the error analysis, we calculate the standard deviation of the AUC metric on the test set through 10 random experiments.
The standard deviation is calculated as follows:
\begin{equation} \label{standard_deviation}
    deviation = \sqrt{\frac{1}{N}\sum_{i=1}^{N}(s_{i} - \frac{1}{N}\sum_{j=1}^{N}s_{j})^2 },
\end{equation}
where $N$ denotes the number of random trials and $s_{i}$ means the AUC score at the $i$-th experiment.

\begin{table*}[htbp]\normalsize
\caption{Results on all the datasets, where the best results are marked in bold and the second best results are underlined. AUC is the evaluation metric.}
\centering
\scalebox{0.8}{
\begin{tabular}{c|ccccccc}
\specialrule{0.1em}{3pt}{3pt}
Method& Syn1 & Syn2 & Syn3 & Coat & Adult & Bank & Credit \\  
\specialrule{0.05em}{3pt}{3pt}
No-selection & 0.5934±0.0001  & 0.8268±0.0004  & 0.7001±0.0001  & 0.6598±0.0172  & 0.8908±0.0043  & 0.9175±0.0004  & 0.8857±0.0003  \\
Lasso & \underline{0.6868±0.0012}  & 0.8651±0.0006  & 0.7199±0.0002  & 0.6482±0.0009  & 0.8972±0.0045  & 0.8887±0.0004  & \underline{0.9460±0.0005}  \\
Tree & 0.6786±0.0002  & 0.8840±0.0003  & 0.7241±0.0001  & 0.6538±0.0176  & 0.8966±0.0041  & 0.8854±0.0006  & 0.9251±0.0221   \\
\specialrule{0.05em}{3pt}{3pt}
L2X & 0.6218±0.0083  & 0.8758±0.0206  & 0.6874±0.0018  & 0.6631±0.0093  & 0.8985±0.0035  & 0.9019±0.0026 & 0.9258±0.0090  \\
CL2X & 0.6262±0.0431  & 0.8278±0.0003  & 0.6941±0.0015  & 0.6627±0.0011  & 0.9046±0.0044  & 0.9160±0.0001  & 0.9122±0.0049  \\
INVASE & 0.6442±0.0011  & 0.8842±0.0002  & 0.7765±0.0009  & 0.6693±0.0115  & 0.8996±0.0036  & 0.9179±0.0835  & 0.9107±0.0370 \\
TabNet & 0.6732±0.0006  & \textbf{0.9068±0.0001}  & \underline{0.7819±0.0002}  & \underline{0.6694±0.0010}  & \underline{0.9074±0.0001}  & \underline{0.9182±0.0022}  & 0.9308±0.0347 \\
\specialrule{0.05em}{3pt}{3pt}
DIWIFT & \textbf{0.6900±0.0019}  & \underline{0.9013±0.0055}  &  \textbf{0.7851±0.0014} & \textbf{0.6736±0.0034} & \textbf{0.9231±0.0031}  & \textbf{0.9231±0.0055} & \textbf{0.9495±0.0019}  \\
\specialrule{0.1em}{3pt}{3pt}
\end{tabular}}
\label{tab:result}
\end{table*}
\subsection{RQ1: Performance Comparison}
To verify the effectiveness of our DIWIFT, we conduct the comparative experiments with all the baselines.
The detailed results are shown in Table~\ref{tab:result}.
The method named ``No-selection'' is a baseline that uses all the features rather than some selected features.
From Table~\ref{tab:result}, we have the following observations: 
1) the comparison results between the baselines for instance-wise feature selection and global feature selection are inconsistent across different datasets. TabNet is the best baseline overall. This may be because L2X and CL2X can only discover a fixed number of influential features per instance, and INVALSE is not specifically designed for tabular data. However, TabNet does not have these limitations.
2) our DIWIFT outperforms all the baselines in most cases, except slightly weaker than TabNet on the Syn2 dataset. In particular, our DIWIFT significantly outperforms TabNet on the syn3 dataset with varying numbers of influential features per instance, as well as on all the four real-world datasets.

\subsection{RQ2: Visual Verification of DIWIFT}
\begin{figure}[htbp]
    \centering
    \includegraphics[width=0.7\linewidth]{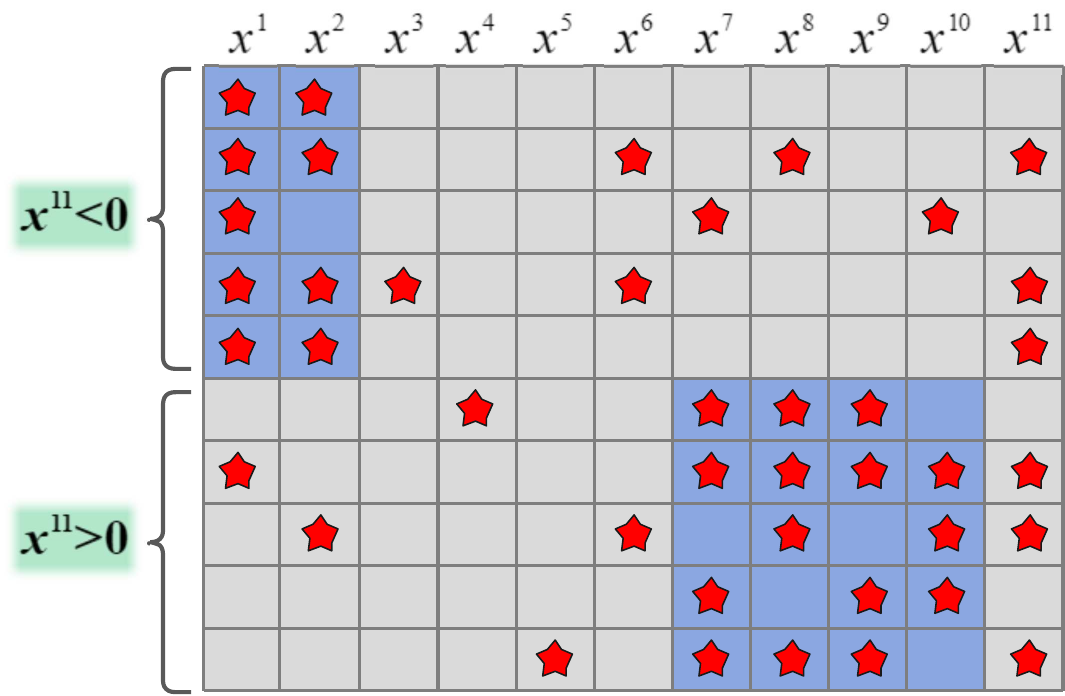}
    \caption{An example of the feature selection results our DIWIFT has on the syn3 dataset, where each row represents an instance, each column represents a feature, the blue squares represent the ground truth of the influential features in each instance, and the stars represent the selected features by our DIWIFT. Note that with $x^{11}$ as a switch feature, the first five instances follow Syn1, and the last five instances follow Syn2.}
    \vspace{-5pt}
    \label{visual}
\end{figure}

Does our DIWIFT effectively discover instance-wise influential features?
To answer this question, we randomly sample 10 instances from the Syn3 dataset, of which 5 instances follow Syn1 and the remaining instances follow Syn2, since the ground truth of the instance-wise influential features is given in the Syn3 dataset.
These 10 instances are then fed into the feature selection module of our DIWIFT to get the feature selection result corresponding to each instance.
The results are shown in Figure~\ref{visual}, where each row represents an instance, each column represents a feature, the blue squares represent the ground truth of influential features in each instance, and the stars represent the feature selection results of our DIWIFT.
We can find that our DIWIFT can identify the most influential instance-wise features and removing the most non-influential features.
This again verifies the validity of our DIWIFT.

\subsection{RQ3: Ablation Study}
\begin{figure}[htbp]
    \vspace{-10pt}
    \centering
    \includegraphics[width=0.9\linewidth]{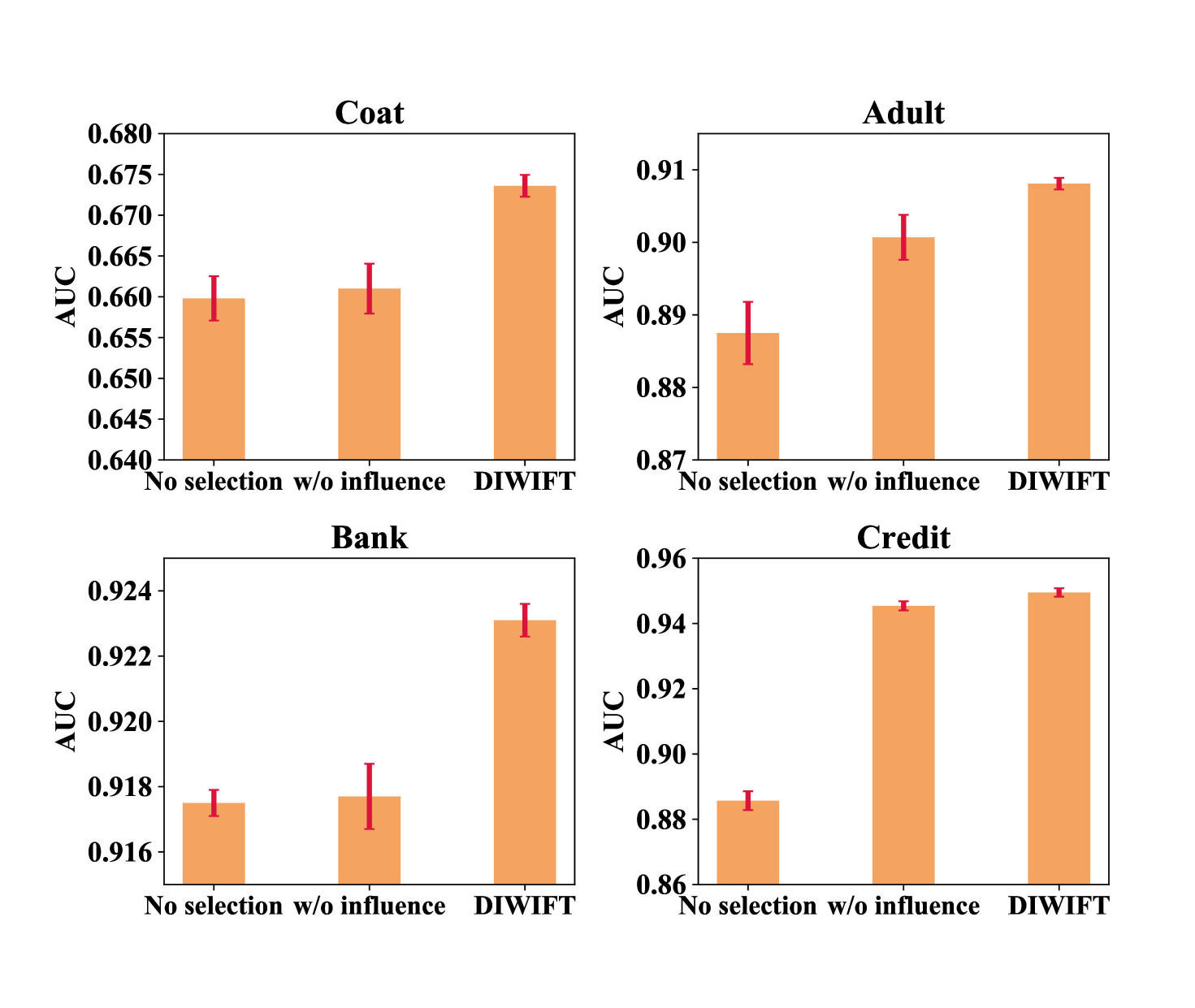}
    \caption{Ablation studies on four real-world datasets.}
    \vspace{-10pt}
    \label{ablation}
\end{figure}

As described in Section~\ref{methodSec}, the feature selection module and the IF calculator are the core modules of our DIWIFT.
To analyze their respective roles, we conduct ablation studies on our DIWIFT using the four real-world datasets.
The results are shown in Figure~\ref{ablation}, where ``w/o influence'' means that only the IF calculator is removed (i.e., a self-attention module trained using a traditional loss function is retained), and ``No-selection'' means that both the feature selection module and the IF calculator are removed.
We can see that removing any module will hurt the performance.
In addition, ``w/o influence'' is weaker than our DIWIFT, but is still better than ``No-selection'', which also shows the importance of feature selection for modeling tabular data.

\subsection{RQ4\&RQ5: In-Depth Analysis of DIWIFT}\label{robust}
\begin{figure}[htbp]
    \centering
    \includegraphics[width=0.7\linewidth]{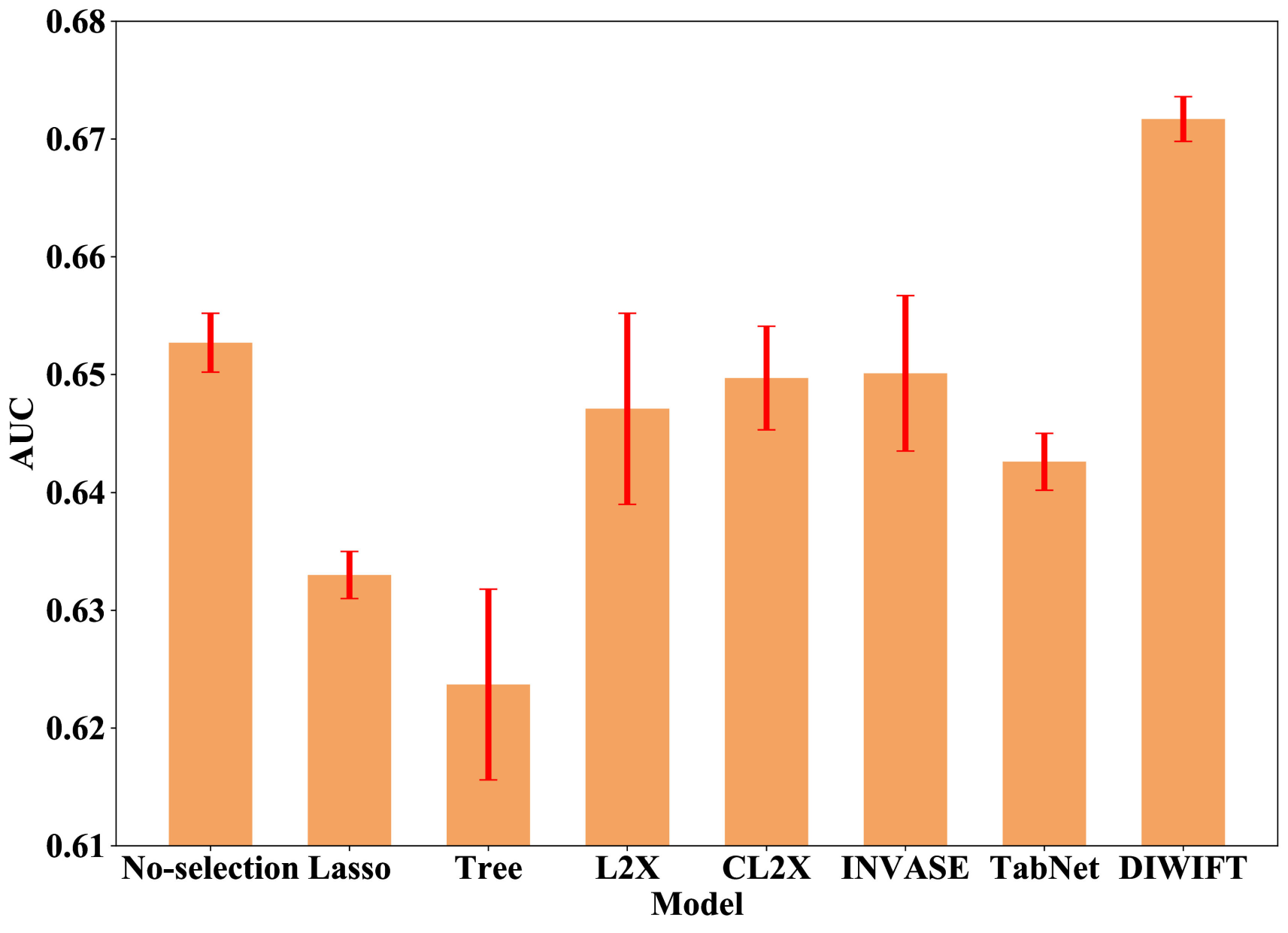}
    \caption{Robustness analysis on Coat.}
    \label{shift}
\end{figure}

As described in Section~\ref{subsec:theoretical}, most existing works on feature selection has not considered the variability of influential features across different scenarios.
Conversely, our DIWIFT can benefit from the influence function to achieve a trade-off between training and validation distributions, i.e., it is relatively more robust.
To evaluate the robustness of all the methods under a distribution shift scenario, we choose the Coat dataset in our experiments.
The Coat dataset contains the sets collected from two sources: one is a biased data collected through the normal user interactions on an online web-shop platform, and the other is an unbiased data collected through a randomized experiment, in which all items that a user can see are randomly assigned by the system.
Clearly, there is a distribution shift between these two sets.
We re-partition Coat, where the set of biased data is used as a training set, and the set of unbiased data is randomly divided into a validation set and a test set with equal size.
We then re-execute and evaluate all the methods using the same hyperparameter search range, and show the results in Figure~\ref{shift}.
Comparing with the Coat column in Table~\ref{tab:result}, we can find that all the baselines have a significant drop in performance, which is even weaker than ``No-selection''.
This shows that the existing feature selection methods are susceptible to distribution shift.
Conversely, our DIWIFT is more robust and has a distinct advantage in different scenarios.

\begin{figure}[htbp]
    \centering
    \includegraphics[width=0.7\linewidth]{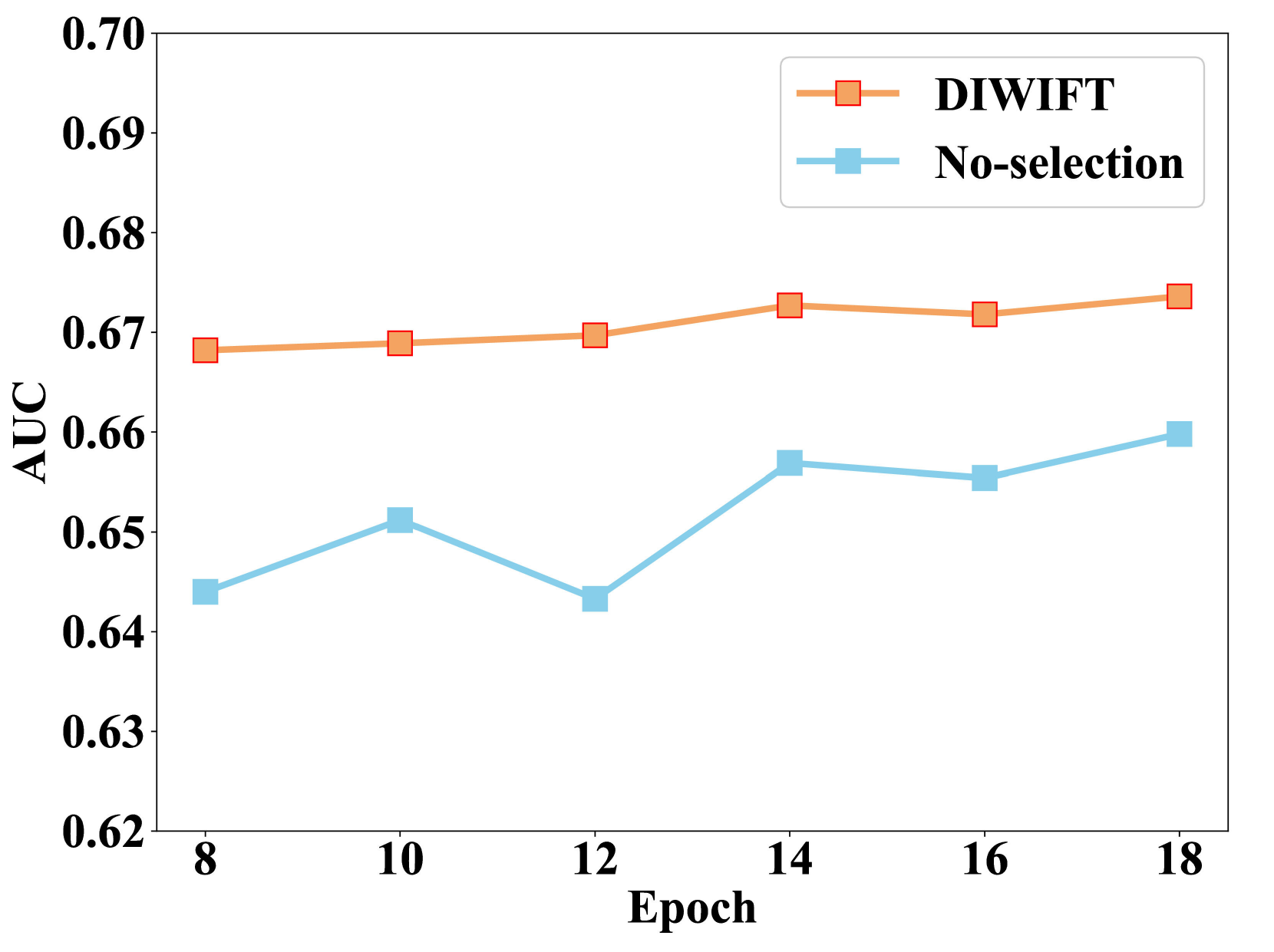}
    \caption{Sensitivity analysis of our DIWIFT with different pre-trained models on Coat.}
    \vspace{-5pt}
    \label{sensitive}
\end{figure}

Finally, we perform a sensitivity analysis of our DIWIFT with the pre-trained models of different performance.
As described in Section~\ref{subsubsec:architecture}, the influence function calculator is an important module of our DIWIFT, and its computation requires the parameters of a pretrained model $\hat{\theta}$.
Since the computed value of the influence function will be used to guide the training of a feature selection module, which is another core module of our DIWIFT, it is necessary to analyze the sensitivity of our DIWIFT on pretrained models with different performance.
Next, we examine this sensitivity of our DIWIFT by conducting a preliminary experiment on Coat.
When obtaining the results of our DIWIFT on Coat as shown in Table~\ref{tab:result}, the optimal number of training iterations for the pre-trained model is 18.
Therefore, we choose the base model when the number of training iterations is 8, 10, 12, 14, and 16 as the pre-trained model, respectively.
We then retrain our DIWIFT and evaluate its performance.
The results are shown in Figure~\ref{sensitive}. 
We can see that our DIWIFT is relatively insensitive to the pre-trained model.
This observation is important for deploying our DIWIFT in a real-world scenario, as it shows that we can use the same pre-trained model for a period of time, which will effectively save training time.

\section{Conclusions}
In this paper, we propose a new perspective based on the influence function for instance-wise feature selection and give some corresponding theoretical insights.
We then propose a new method for discovering instance-wise influential features in tabular data (DIWIFT), where a feature selection module with a self-attention network is used to compute the selection probabilities of all features from each instance, and an influence function calculator is used to calculate the corresponding influence and guide the feature selection module through a back propagation.
We conduct extensive experiments on some synthetic and real-world datasets, where the results validate the effectiveness and robustness of our DIWIFT.

\begin{acks}
We thank the support of National Natural Science Foundation of China Nos. 61836005, 62272315 and 62172283.
\end{acks}

\bibliographystyle{ACM-Reference-Format}
\bibliography{sample-base}


\appendix
\section{Appendix}\label{appendix}
\begin{algorithm}[htpb]
	\caption{Calculating the influence function $\phi_i(p(\bm{x},\omega))$}
	\label{alg:IF}
	\begin{algorithmic}[1]
		\REQUIRE The training set after feature reweighting $\{p_i\odot \bm{x}_i,y_i\}_{i=1}^n$, validation set $\{\{p_j\odot \bm{x}_j,y_j\}_{j=1}^m$, pre-trained base network $\hat{\theta}$.
		\STATE Calculate the gradient of the validation loss, i.e., $\mu=[\nabla_{\theta}\sum_{j=1}^ml(p_j\odot\bm{x}_j,y_j,\hat{\theta})]$.
		\STATE Initialize $H_0^{-1}\mu=\mu$.
		\STATE \textbf{repeat to get HVP}
		\STATE Uniformly sample some instances from the training set to calculate the estimated Hession matrix $\tilde{H}$.
		\STATE Calculate $H_u^{-1}\mu=\mu+(I-\tilde{H})H_{u-1}^{-1}\mu$.
		\STATE \textbf{until} convergence
		\STATE Refer to $h$ as the converged HVP, which is the estimation of $H_{\hat{\theta}}(p)^{-1}[\nabla_{\theta}\sum_{j=1}^ml(p_j\odot\bm{x}_j,y_j,\hat{\theta})]$.
		\STATE Calculate $\nabla_{\theta}l(p_i\odot\bm{x}_i,y_i,\hat{\theta})$.
		\STATE Calculate the gradient of $h^T\nabla_{\theta}l(p_i\odot\bm{x}_i,y_i,\hat{\theta})\in\mathbb{R}$ over $\bm{x}$.
	\end{algorithmic}
\end{algorithm}

\newpage
\begin{algorithm}[htpb]
	\caption{Discovering Instance-wise Influential Features in Tabular Data (DIWIFT)}
	\label{alg:diff}
	\begin{algorithmic}[1]
		\REQUIRE The training set $\{z_i\}_{i=1}^n$, validation set $\{z_j\}_{j=1}^m$.
		\STATE Train a base model to get $\hat{\theta}$.
		\STATE Initialize all parameters of a feature selection model with a self-attention network.
		\STATE \textbf{repeat}
		\STATE Fed instances into feature selection model to get probability $p_i$ and $p_j$ using Eq.\eqref{prob}.
		\STATE Calculate the influence function in Eq.\eqref{phip} using the stochastic estimation method.
		\STATE Calculate the loss in Eq.\eqref{loss}.
		\STATE Do back-propagation to update the parameters of the feature selection model.
		\STATE \textbf{until} convergence
	\end{algorithmic}
\end{algorithm}

\end{document}